\documentclass{ecai}
\usepackage{times}
\usepackage{graphicx}
\usepackage{latexsym}
\usepackage{amsfonts, amsopn, amssymb, epsfig}
\usepackage[cmex10]{amsmath}
\usepackage{algorithm}
\usepackage[noend]{algorithmic}

\begin{document}

\title{Empirical study of PROXTONE and PROXTONE$^+$ for Fast Learning of Large Scale Sparse Models}

\author{Ziqiang Shi \and Rujie Liu\institute{Fujitsu Research \& Development Center,
China, email: \{shiziqiang, rjliu\}@cn.fujitsu.com} }

\maketitle
\bibliographystyle{ecai}

\begin{abstract}
  PROXTONE is a novel and fast method for optimization of large scale non-smooth convex problem~\cite{shi2015large}. In this work, we try to use PROXTONE method in solving large scale \emph{non-smooth non-convex} problems, for example training of sparse deep neural network (sparse DNN) or sparse convolutional neural network (sparse CNN) for embedded or mobile device. PROXTONE converges much faster than first order methods, while first order method is easy in deriving and controlling the sparseness of the solutions. Thus in some applications, in order to train sparse models fast,  we propose to combine the merits of both methods, that is we use PROXTONE in the first several epochs to reach the neighborhood of an optimal solution, and then use the first order method to explore the possibility of sparsity in the following training. We call such method PROXTONE plus (PROXTONE$^+$). Both PROXTONE and PROXTONE$^+$ are tested in our experiments, and which demonstrate both methods improved convergence speed twice as fast at least on diverse sparse model learning problems, and at the same time reduce the size to 0.5\% for DNN models. The source of all the algorithms is available upon request.
\end{abstract}

\section{INTRODUCTION}
Benefited from the advances in deep learning and big data, the accuracy has been dramatically improved on difficult pattern recognition problems in vision and speech~\cite{krizhevsky2012imagenet,hinton2012deep}. But currently there are two urgent problems need to solve for real life, especially internet applications of deep learning: the first one is that it always took a very long time to adjust the structures and parameters to obtain a satisfactory deep model; and the second one is how to program the always really big deep network on embedded devices or mobile devices. Thus fast learning of sparse regularized models, for example, such as L1 regularized logistic regression, L1 regularized deep neural network (sparse DNN) or L1 regularized convolutional neural network (sparse CNN) becomes very important.

In order to solve the problem of learning large scale L1 regularized model:
\begin{align}
  \min_{x \in \mathbb{R}^p} \,f(x) := \frac{1}{n}\sum_{i=1}^n g_i (x) + \lambda_2 \|x\|_1,
  \label{eq:composite-form-online}
\end{align}
researchers have proposed the standard and popular \emph{proximal stochastic gradient descent}  methods (ProxSGD), whose main appealing
is that they have an iteration cost which is independent of $n$,
making them suited for modern problems where $n$ may be very large. The basic ProxSGD method for
optimizing~\eqref{eq:composite-form-online}, uses iterations of the form
\begin{equation}\label{eqn:prox-sg}
x^k = \mathcal{S}_{\alpha_k h}[x^{k-1} - \alpha_k \nabla g_{i_k}(x^{k-1}) ] ,
\end{equation}
where $\mathcal{S}_{\varepsilon}[\cdot] $ is the soft-thresholding operator:
\begin{equation}
\mathcal{S}_{\varepsilon}[x]\doteq \left\{ \begin{array}{l}
 x-\varepsilon, \quad \textrm{if } x>\varepsilon   \\
 x+\varepsilon, \quad \textrm{if } x<-\varepsilon,   \\
0,\quad \textrm{otherwise}  \\
 \end{array} \right.
\end{equation}
and at each iteration an index $i_k$ is sampled uniformly from the set $\{1, ..., n\}$. The randomly
chosen gradient $\nabla g_{i_k}(x_{k-1})$ yields an unbiased estimate of the true gradient $\nabla g(x_{k-1})$ and one can show
under standard assumptions that,  for a suitably chosen decreasing
step-size sequence $\{\alpha_k\}$, the ProxSGD iterations have an expected sub-optimality for convex objectives of~\cite{bertsekas2011incremental}
\[
\mathbb{E}[f(x^k)] - f(x^\ast) = O(\frac{1}{\sqrt{k}})£¬
\]
and an expected sub-optimality for strongly-convex objectives of
\[
\mathbb{E}[f(x^k)] - f(x^\ast) = O(\frac{1}{k}).
\]
In these rates, the expectations are taken with respect to the selection of the $i_k$ variables.

Thus at least in theory, in fact also in practice it is showed that ProxSGD is very slow in solving the problem~\ref{eq:composite-form-online}. While in real life applications, we need to learn and adjust fast in order to obtain a usable model quickly. This requirement results in a large variety of approaches available to accelerate the convergence of ProxSGD methods, and a full review of this immense literature would be outside the scope of this work. Several recent work considered various special or general cases
of~\eqref{eq:composite-form-online},
and developed algorithms that enjoy the linear convergence rate, such as ProxSDCA~\cite{shalev2012proximal}, MISO~\cite{mairal2013optimization}, SAG~\cite{schmidt2013minimizing}, ProxSVRG~\cite{xiao2014proximal}, SFO~\cite{sohl2014fast}, ProxN~\cite{lee2012proximal}, and PROXTONE~\cite{shi2015large}. All these methods converge with an exponential rate in the value of the objective function, except that the ProxN achieves superlinear rates of convergence for the \emph{solution}, however it is a batch mode method. Shalev-Shwartz and Zhang's ProxSDCA~\cite{shalev2013stochastic,shalev2012proximal} considered the case where the component functions have the form
$g_i(x)=\phi_i(a_i^T x)$ and the Fenchel conjugate functions of~$\phi_i$ can be computed efficiently. Schimidt et al.'s SAG~\cite{schmidt2013minimizing} and Jascha et al.'s SFO~\cite{sohl2014fast} considered the case where $\lambda_2\equiv 0$.

In order to solve the problem~\eqref{eq:composite-form-online} with linear convergent rate,
we has proposed a novel and fast method called \textbf{prox}imal s\textbf{to}chastic \textbf{N}ewton-type gradient descent (PROXTONE)~\cite{shi2015large}. Compared to previous methods, the PROXTONE like other typical quasi-Newton techniques, requires no adjustment of hyperparameters.
And at the same time, the PROXTONE method has the low iteration cost as that of ProxSGD methods, but achieves
the following convergence rates according to the two theorems in~\cite{shi2015large}
\begin{align}
   \mathbb{E}[f(x^k)] - f^* = O (\mu^k\|x^*-x^0\|^2).
\label{eq:proxtone-value-linear}
\end{align}
When some additional conditions are satisfied, for example $\nabla^2g_i$ are Lipschitz continuous and so on, then PROXTONE converges exponentially to $x^\star$ in expectation
\begin{align*}
  \mathbb{E}[ \|x^{k+1} - x^\star\|] = O(\eta^{k} \|x^*-x^0\|^2).
\end{align*}
For details and proofs, please refer to our previous theory work~\cite{shi2015large}.

The PROXTONE iterations take the form $x^{k+1} \leftarrow x^k + t_k\Delta x^k$, where $\Delta x^k$ is obtained by
\begin{equation}
\label{eqn:proxtone}
    \Delta x^k \leftarrow \arg\min_{d} d^T(\nabla_k + H_kx^k)+\frac{1}{2}d^T H_k d+\lambda_2 \|x^k+d\|_1,
\end{equation}
here $\nabla_{k} = \frac{1}{n}\sum_{i=1}^n \nabla_{k}^i$, $H_{k} =  \frac{1}{n}\sum_{i=1}^n H_{k}^i$, and at each iteration a random index $j$ and corresponding $H_{k+1}^j$ is selected, then we set
\[
\nabla_{k+1}^i = \begin{cases}
 \nabla g_i(x^{k+1})-H_{k+1}^i x^{k+1} & \textrm{if $i = j$,}\\
\nabla_{k+1}^i & \textrm{otherwise.}
\end{cases}
\]
and $H_{k+1}^i\leftarrow H_{k}^i$ ($i\neq j$).

In this work, we try to use the second order method PROXTONE to promote the training of sparse deep models. Compared to conventional methods, PROXTONE can make full use of the gradients, thus needs less gradients (epochs) to achieve same performance, which means converges much fast in the number of epochs. But for each gradient, PROXTONE needs to update the hessian, to construct the low-dimensional space, and solve some kind of lasso subproblem, thus needs much more CPU time against first order methods. That means, finally PROXTONE may converges slow in time than first order methods. In order to overcome this problem, in each iteration, we performance less iterations in solving the subproblems, which means we are satisfied with less exact steepest search directions. This approximation accelerate the convergence of PROXTONE, but result in less sparsity in weights of deep neural networks.

During the empirical study, we found that in some situations, for example training of fully connected DNN, fast approximated PROXTONE cannot fully explore the possibility of sparseness in weights. While first order method is easy in deriving and accumulating the sparseness in each iteration by soft threshold operators, thus we propose to combine first order method with PROXTONE in training DNN. We call such kind of methods PROXTONE$^+$. Experiments show that PROXTONE and PROXTONE$^+$ are suitable for training different kind of neural networks, for example PROXTONE is much suitable for sparse CNN, since whose almost all weights are of shared type, while PROXTONE$^+$ is much more suitable for training of sparse DNN.
Finally, the optimizer and the code (matlab and python) reproduce the figures in this work is available upon request.

We now outline the rest of this study. Section~\ref{sec:ALGORITHM} presents the main PROXTONE algorithm for L1 regularized model learning, and states choice and details in the implementation. Section~\ref{sec:PROXTONEplus} describe the PROXTONE$^+$ method. We report some experimental results in
Section~\ref{sec:EXPERIMENTAL}, and provide concluding remarks in Section~\ref{sec:CONCLUSION}.

\section{ALGORITHM}
\label{sec:ALGORITHM}

Our goal is to use the PROXTONE for sparse regularized model learning. In general, we always separate the $n$ training samples into $M$, for example several hundred mini-batches, but in order for the simplicity of notations and description, we did not distinguish between $n$ and $M$. That is in the following algorithms, $n$ means the number of mini-batches, which should be keep in mind. In this section, we first describe the general procedure by which we optimize the parameter $x$. We then describe the procedure of the BFGS~\cite{nocedal2006numerical} method by which the online Hessian approximation is maintained for each batch or subfunction. This followed by a description of solving the subproblem in PROXTONE.

\subsection{PROXTONE}

In each iteration, general PROXTONE uses a L1 regularized piecewise quadratic function to approximate the target loss function for the deep model in a local area around the current point $x^{k+1}$, and the solution of the regularized quadratic model is used to be the new point. The component function $g_{i_k}(x)$ is sampled randomly, and then the gradient and the approximation of the hessian is used to update the the regularized quadratic model. The procedure is summarized in the Algorithm~\ref{alg:sparse-proxtone}.

\begin{algorithm}
\caption{PROXTONE for L1 regularized model learning}
\label{alg:sparse-proxtone}

\textbf{Input}: start point $x^0 \in$ dom $f$; for $i\in\{1,2,..,n\}$, let $H_{-1}^i=H_0^i$ be a positive definite approximation to the Hessian of $g_i(x)$ at $x^0$, $\nabla_{-1}^i=\nabla_0^i=\nabla g_i(x^0) - H_0^ix^0$, and let $g_i^0(x)=g_i(x^0)+(x-x^0)^T\nabla g_i(x^0)+\frac{1}{2}(x-x^0)^TH_0^i(x-x^0)$; $G^0(x)=\frac{1}{n}\sum_{i=1}^n g_i^0(x)$; $y\in \mathbb{R}^{p*\text{MAX\_HISTORY}*n}$, the history of gradient changes for all $i\in\{1,2,..,n\}$; $last\_x\in \mathbb{R}^{p*n}$ and $last\_df\in \mathbb{R}^{p*n}$ holds the last position and the last gradient for all the objective functions; MAX\_HISTORY; $s\in \mathbb{R}^{p*\text{MAX\_HISTORY}*n}$, the history of $x$ or position changes for all $i\in\{1,2,..,n\}$.

1: \textbf{repeat}

2: Solve the subproblem (it is indeed the well known lasso problem) for new approximation of the solution:
\begin{align}
x^{k+1} \leftarrow \arg\min_{x} \bigl[ G^k(x) + \lambda_2 \|x\|_1 \bigr] .
  \label{eq:search-direction}
\end{align}

3: Sample $i_k$ from $\{1,2,..,n\}$, update the history of position and gradient differences for the mini-batch $i_k$:
\begin{align}
s(:,2:\text{MAX\_HISTORY},i_k) &= s(:,1:\text{MAX\_HISTORY}-1,i_k)\nonumber\\
s(:,1,i_k) &= x - last\_x(:,i_k)\nonumber \nonumber\\
y(:,2:\text{MAX\_HISTORY},i_k) &= y(:,1:\text{MAX\_HISTORY}-1,i_k)\nonumber\\
y(:,1,i_k) &= \nabla g_{i_k}(x^{k+1}) - last\_df(:,i_k)\nonumber
\end{align}

4: Update the Hessian approximation $H_{k+1}^{i_k}$ for the mini-batch $i_k$ (described in detail in Algorithm~\ref{alg:hessian_update});

5: Update the quadratic models or surrogate functions:
\begin{align}
g_{i_k}^{k+1}(x)&=g_{i_k}(x^{k+1})+(x-x^{k+1})^T\nabla g_{i_k}(x^{k+1}) \nonumber \\
&+\frac{1}{2}(x-x^{k+1})^TH_{k+1}^{i_k}(x-x^{k+1}),
  \label{eq:subfunction-surrogate-update}
\end{align}
while leaving all other $g_i^{k+1}(x)$ unchanged: $g_i^{k+1}(x)\leftarrow g_i^{k}(x)$ ($i\neq j$); and $G^{k+1}(x)=\frac{1}{n}\sum_{i=1}^n g_i^{k+1}(x)$.

6: \textbf{until} stopping conditions are satisfied.

\textbf{Output}: $x^k$.
\end{algorithm}

In deep learning, the dimensionality of $x$ is always large. As a result, the memory and computational cost of working directly with the matrices in Algorithm~\ref{alg:sparse-proxtone} is prohibitive, as is the cost of storing the history terms and required by BFGS. Thus we employ the idea from~\cite{sohl2014fast}, that is we construct a shared low dimensional subspace which makes the algorithm tractable in terms of computational overhead and memory for large problems. $x$ and the gradients are mapped into a limited sized shared adaptive low-dimensional space, which is expanded when meeting a new observation. The Hessian, the regularized quadratic model, and further the solution are updated in this low-dimensional space. Finally then solution is projected back to the original space to become the real optimal points. This mapping or projection is comprised of a dense matrix, thus the sparse solution in low-dimensional space may result in non-sparse solution in original space. This problem will be discussed and solved in Section~\ref{sec:PROXTONEplus}.

\subsection{Hessian approximation}

Arguably, the most important feature of this method is the regularized quadratic model, which incorporates second order information in the form of a positive definite matrix $H_k^{i_k}$. This is key because, at each iteration, the
user has complete freedom over the choice of $H_k^{i_k}$. A few suggestions for the choice of $H_k^i$ include: the simplest option is $H_k^i=I$ that no second order information is employed; $H_k^{i_k}=\nabla^2g_i(x_k)$ provides the most accurate second order information,
but it is (potentially) much more computationally expensive to work with; in order to do a tradeoff between accuracy and complexity, the most popular formulae for updating the Hessian approximation is the BFGS formula, which is defined by
\begin{align}\label{eqn:bfgs}
    B_k = B_{k-1}-\frac{B_{k-1}s_{k-1}s^T_{k-1}B_{k-1}}{s^T_{k-1}B_{k-1}s_{k-1}}+\frac{y_{k-1}y_{k-1}^T}{y_{k-1}^Ts_{k-1}},
\end{align}
where
\[
y_{k-1} = \nabla f(x_k) -  \nabla f(x_{k-1}), \quad s_{k-1} = x_k - x_{k-1}.
\]
We store a certain number (say, MAX\_HISTORY) of the vector pairs $\{s_k, y_k\}$ used in the above formulas.
After the new iteration is computed, the oldest vector pair in the set of pairs $\{s_i, y_i\}$ is replaced by the new pair $\{s_k, y_k\}$ obtained from the above step. In this way, the set of vector pairs includes curvature information from the MAX\_HISTORY most recent iterations. This is indeed the famous limited-memory BFGS algorithm, which can be stated formally as the following Algorithm~\ref{alg:hessian_update}.

\begin{algorithm}[H]
\caption{Update the Hessian approximation for the mini-batch $i_k$}
\label{alg:hessian_update}

\textbf{Input}: MAX\_HISTORY = 20, $s\in \mathbb{R}^{p*\text{MAX\_HISTORY}*n}$, the history of $x$ or position changes for all $i\in\{1,2,..,n\}$, $y\in \mathbb{R}^{p*\text{MAX\_HISTORY}*n}$, the history of gradient changes for all $i\in\{1,2,..,n\}$, $H_{k+1}^{i_k}=I$.

1: \textbf{for} $j =1:\text{MAX\_ITER}$

2: \quad $tmpy = y(:,\text{MAX\_HISTORY} + 1 - j,i_k)$;

3: \quad $tmps = s(:,\text{MAX\_HISTORY} + 1 - j,i_k)$;

4: \quad \textbf{if } $tmpy^T * tmps > 0$

5: \quad \quad  $H_{k+1}^{i_k} = H_{k+1}^{i_k} - (H_{k+1}^{i_k} * (tmps * tmps^T) *H_{k+1}^{i_k}) / (tmps^T * H_{k+1}^{i_k} * tmps)  + (tmpy * tmpy^T)/ (tmpy^T * tmps)$;

6:  \quad \textbf{end if}

7: \textbf{end for}

\textbf{Output}: $H_{k+1}^{i_k}$.
\end{algorithm}

After the obtaining of $H_{k+1}^{i_k}$, then we can update the local regularized quadratic model (the subproblem), which can be solved by a proximal algorithm.

\subsection{The subproblem}

The subproblem~(\ref{eq:search-direction}) is a lasso problem, which can be effectively and accurately solved by the proximal algorithms~\cite{parikh2013proximal}. It is summarized in Algorithm~\ref{alg:prox_lasso}.

\begin{align}
x^{k+1} \leftarrow \arg\min_{x} \bigl[ G^k(x) + \lambda_2\|x\|_1 \bigr] = \arg\min_{x} F^k(x).
  \label{eq:search-direction-l1lr}
\end{align}

\begin{algorithm}[H]
\caption{Solving subproblem~(\ref{eq:search-direction}) based on proximal algorithms}
\label{alg:prox_lasso}

\textbf{Input}: start point $x_0 = x^k$, MAX\_ITER = 100, ABSTOL   = 1E-5, $\lambda$ = 1, $\beta$ = 0.5.

1: \textbf{for} i=1:MAX\_ITER

2: \quad grad\_x = $\nabla G^k(x_{i-1})$,

3: \quad \textbf{while} 1

4: \quad \quad $z = \mathcal{S}_{\lambda*\lambda_2}[x_{i-1} - \text{lambda}*\text{grad\_x}]$,

5: \quad \quad  \textbf{if}$ G^k(z) <= G^k(x_{i-1}) + \text{grad\_x}'*(z - x_{i-1}) + (1/(2*\lambda))*\|z - x_{i-1}\|^2$

6:  \quad \quad \quad break;

7:  \quad \quad \textbf{end if}

8:  \quad \quad $\lambda = \beta*\lambda$;

9: \quad \textbf{end while}

10: \quad  $x_i = z$;

11:    \quad    \textbf{if} $i > 1$ \&\& $|F^k(x_i) -  F^k(x_{i-1})| < \text{ABSTOL}$

12:      \quad \quad     break;

13:    \quad    \textbf{end if}

14: \textbf{end for}

\textbf{Output}: $x^{k+1}=x_i$.
\end{algorithm}

That means for each gradient, we need to use several iterations of computing approximated Hessian to forming a lasso problem, which also needs several iterations to solve. Thus typically PROXTONE needs much more time for each iteration than that of first order method. That means although PROXTONE is much fast than other other methods in the number of gradients or epochs, but may be slower in time. In the following section, we will try to solve this problem.

\section{The PROXTONE$^+$}
\label{sec:PROXTONEplus}

Compared to conventional method, PROXTONE can achieve the same performance with less gradients, that is in less epochs. But since it always needs much more computation than first order method for each iteration, thus always PROXTONE converges slowly than first order methods in physic time. In order to speed up the PROXTONE, we try to not solve the lasso problem so exactly, that is we always set 'MAX\_ITER = 1' in the Algorithm~\ref{alg:prox_lasso}. This result in inexact solution in each iteration of PROXTONE, but also result in much faster convergence speed. This speed up cause new problems, that is we cannot control the sparseness of the solution. In order to overcome this problem, we try to combine PROXTONE with first order method, that is in the first stage, we use PROXTONE to reach the nearby of the optimal, and then comes to the second stage, we use ProxSAG to further explore the possibility of sparseness of the solution. The rough idea result in the following PROXTONE$^+$ algorithms.

\begin{algorithm}
\caption{PROXTONE$^+$ for L1 regularized model learning}
\label{alg:sparse-proxtone-plus}

\textbf{Input}: start point $x^0 \in$ dom $f$; for $i\in\{1,2,..,n\}$, let $H_{-1}^i=H_0^i$ be a positive definite approximation to the Hessian of $g_i(x)$ at $x^0$, $\nabla_{-1}^i=\nabla_0^i=\nabla g_i(x^0) - H_0^ix^0$, and let $g_i^0(x)=g_i(x^0)+(x-x^0)^T\nabla g_i(x^0)+\frac{1}{2}(x-x^0)^TH_0^i(x-x^0)$; and $G^0(x)=\frac{1}{n}\sum_{i=1}^n g_i^0(x)$; $N$, the number of epochs to perform PROXTONE.

1: \textbf{repeat}

2: \textbf{if} $k < N$ (use PROXTONE)

3: \quad Solve the lasso subproblem for new approximation of the solution:
\begin{align}
x^{k+1} \leftarrow \arg\min_{x} \bigl[ G^k(x) + \lambda_2 \|x\|_1 \bigr] .
  \label{eq:search-direction}
\end{align}

4: \quad Sample $i_k$ from $\{1,2,..,n\}$, and update the quadratic models or surrogate functions:
\begin{align}
g_{i_k}^{k+1}(x)&=g_{i_k}(x^{k+1})+(x-x^{k+1})^T\nabla g_{i_k}(x^{k+1}) \nonumber \\
&+\frac{1}{2}(x-x^{k+1})^TH_{k+1}^i(x-x^{k+1}),
  \label{eq:subfunction-surrogate-update}
\end{align}
while leaving all other $g_i^{k+1}(x)$ unchanged: $g_i^{k+1}(x)\leftarrow g_i^{k}(x)$ ($i\neq i_k$); and $G^{k+1}(x)=\frac{1}{n}\sum_{i=1}^n g_i^{k+1}(x)$.

5: \textbf{else} (use ProxSAG)

6: \quad Sample $i_k$ from $\{1,2,..,n\}$, and update the gradient $y_{k,i}$ and the average gradient $y_{k-1}$:
\begin{equation}
\label{eq:yi}
y_{k,i} = \begin{cases}
\nabla g_i(x_k) & \textrm{if $i = i_k$,}\\
y_{k-1,i} & \textrm{otherwise.}
\end{cases}
\end{equation}
\begin{align}
y_k=\frac{1}{n}\sum_{i=1}^n y_{k,i}
\end{align}
and finally the update of $x^{k+1}$:
\begin{align}
x^{k+1} = \mathcal{S}_{\lambda_2 / L}[x^k -  y_k / L ] ,
\end{align}

7: \textbf{end if}

8: \textbf{until} stopping conditions are satisfied.

\textbf{Output}: $x^k$.
\end{algorithm}

\section{EXPERIMENTAL RESULTS}
\label{sec:EXPERIMENTAL}

We compared our optimization technique to several competing optimization techniques for several objective functions.
The results are illustrated in Figures~\ref{fig results},~\ref{fig results_sparse_dnn}, and~\ref{fig results_sparse_cnn}, and the optimization techniques and objectives are described below.
For all problems our method outperformed all other techniques in the comparison.

\subsection{Sparse regularized logistic regression}
In our preliminary study, we use some large scale convex problems to debug our algorithm. Here present the results of some numerical experiments to illustrate the properties of the PROXTONE method. We focus on the sparse regularized logistic regression problem for binary classification: given a set of training examples
$(a_1,b_1),\ldots, (a_n, b_n)$ where $a_i\in\mathbb{R}^p$ and $b_i\in\{+1, -1\}$, we find the optimal predictor $x\in\mathbb{R}^p$ by solving
\[
 \min_{x\in\mathbb{R}^p} \quad
    \frac{1}{n} \sum_{i=1}^n \log\bigl(1+\exp(-b_i a_i^T x)\bigr)
    + \lambda_1 \|x\|_2^2 + \lambda_2 \|x\|_1 ,
\]
where $\lambda_1$ and $\lambda_2$ are two regularization parameters.
We set
\begin{equation}\label{eqn:L2inF}
    g_i(x)=\log(1+\exp(-b_i a_i^T x)  + \lambda_1\|x\|_2^2, \qquad
    h(x)=\lambda_2\|x\|_1,
\end{equation}
and
\[
 \lambda_1=1E-4, \qquad \lambda_2=1E-4.
\]

We used some publicly available data sets. The \emph{protein} data set was obtained from the KDD Cup 2004\footnote{http://osmot.cs.cornell.edu/kddcup}; the covertype data sets were obtained from the LIBSVM Data\footnote{http://www.csie.ntu.edu.tw/$\thicksim$cjlin/libsvmtools/datasets}.

The performance of PROXTONE is compared with some related algorithms:
\begin{itemize}\itemsep 0pt
    \item ProxSGD (Algorithm~\ref{alg:sparse-sgd}): We used a constant step size that gave the best performance among
        all powers of~$10$;
    \item ProxSAG (Algorithm~\ref{alg:sparse-sag}): This is a proximal version of the SAG method, with the trailing number providing
the Lipschitz constant;
\end{itemize}

\begin{algorithm}
\caption{ProxSGD for L1 regularized model learning}
\label{alg:sparse-sgd}

\textbf{Input}: start point $x^0 \in$ dom $f$.

1: \textbf{repeat}

2: Sample $i_k$ from $\{1,2,..,n\}$,
\begin{align}
x^{k+1} = \mathcal{S}_{\eta \lambda_2}[x^k - \eta \nabla g_{i_k}(x^k) ] ,
\end{align}

3: \textbf{until} stopping conditions are satisfied.

\textbf{Output}: $x^k$.
\end{algorithm}

\begin{algorithm}
\caption{ProxSAG for L1 regularized model learning}
\label{alg:sparse-sag}

\textbf{Input}: start point $x^0 \in$ dom $f$; let $y_{0,i}=\nabla g_i(x^0)$, and $y_0=\frac{1}{n}\sum_{i=1}^n y_{0,i}$ be the average gradient.

1: \textbf{repeat}

2: Sample $i_k$ from $\{1,2,..,n\}$, and update the gradient $y_{k,i}$ and the average gradient $y_{k-1}$:
\begin{equation}
\label{eq:yi}
y_{k,i} = \begin{cases}
\nabla g_i(x_k) & \textrm{if $i = i_k$,}\\
y_{k-1,i} & \textrm{otherwise.}
\end{cases}
\end{equation}
\begin{align}
y_k=\frac{1}{n}\sum_{i=1}^n y_{k,i}
\end{align}
and finally the update of $x^{k+1}$:
\begin{align}
x^{k+1} = \mathcal{S}_{\lambda_2 / L}[x^k -  y_k / L ] ,
\end{align}

4: \textbf{until} stopping conditions are satisfied.

\textbf{Output}: $x^k$.
\end{algorithm}

\begin{figure}
\centering
\begin{tabular}{ccc}
\hspace{-5mm}
\begin{tabular}{c}
(a)\includegraphics[width=0.39\linewidth]{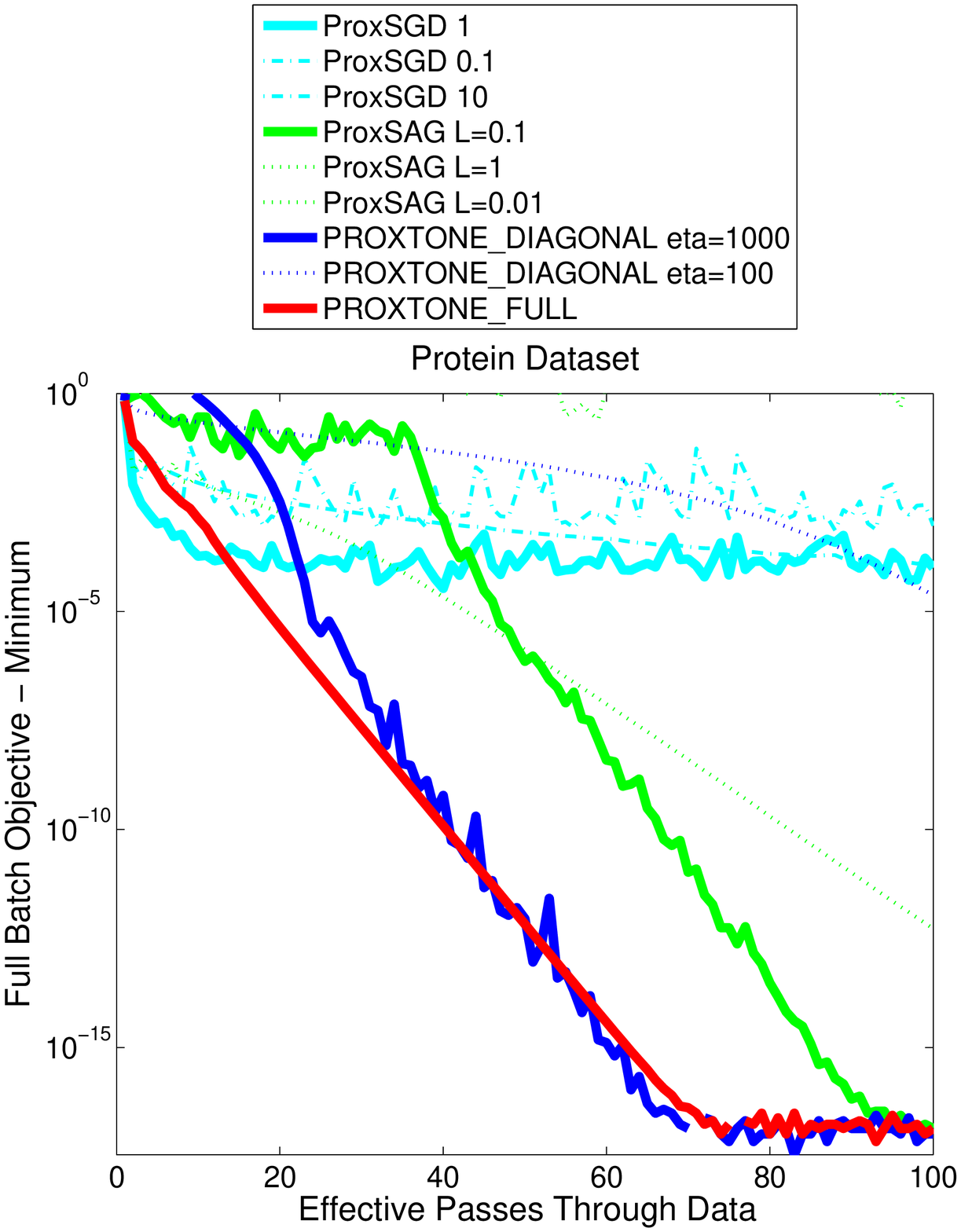}
\end{tabular}
\hspace{-5mm}
 &
\begin{tabular}{c}
(b)\includegraphics[width=0.39\linewidth]{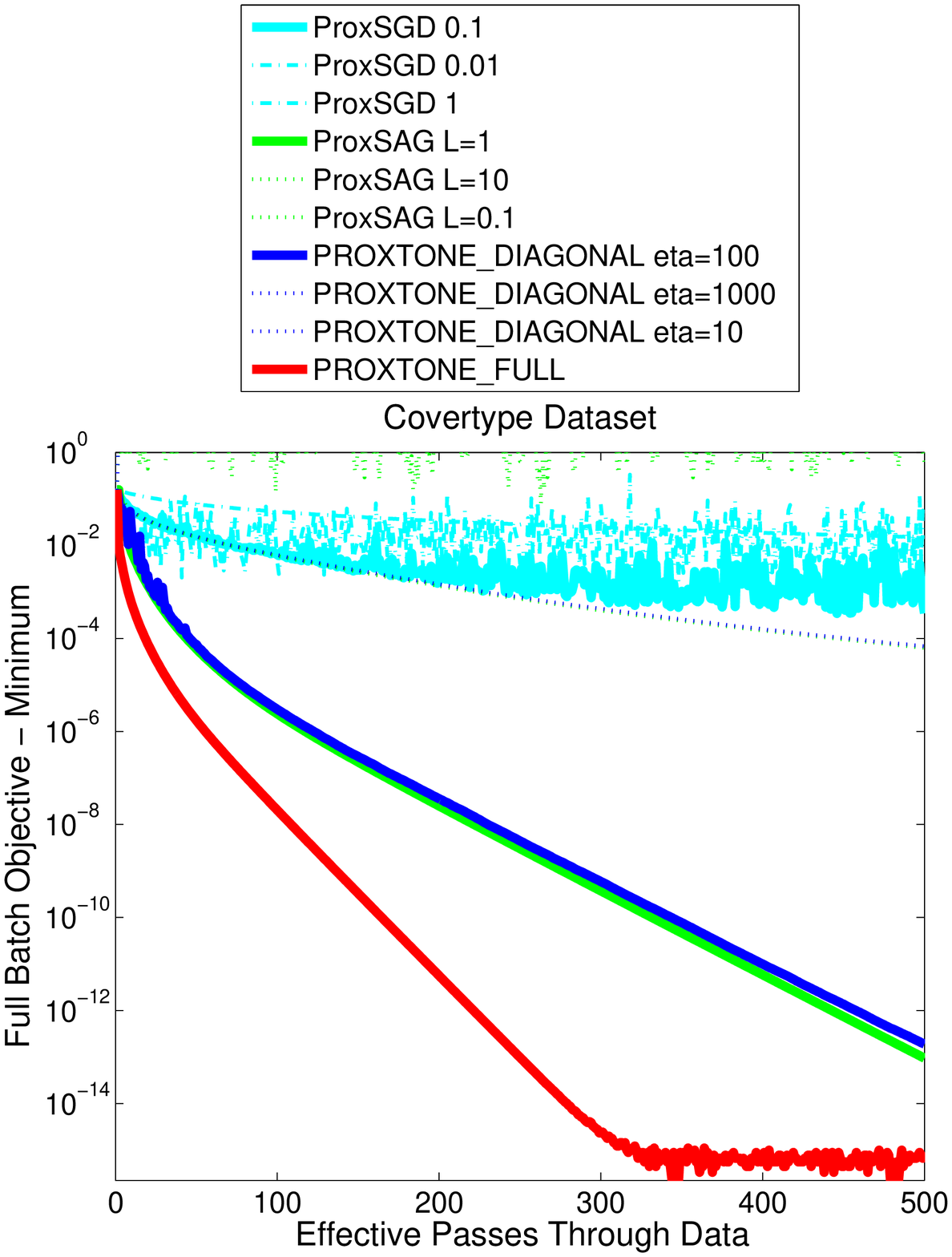}
\end{tabular}
\hspace{-5mm}
\end{tabular}
\caption{
A comparison of PROXTONE to competing optimization techniques in solving sparse regularized logistic regression for two datasets, (a) is protein; (b) is covertype.
The bold lines indicate the best performing hyperparameter for each optimizer.
}
\label{fig results}
\end{figure}

The results of the different methods are plotted for the first 100 and 500 effective passes for protein and covertype respectively through the data in Figure~\ref{fig results}. Here we test PROXTONE with two kinds of Hessian, the first is with diagonal Hessian with constant diagonal elements, and the Hessian of the other kind is updated by Algorithm~\ref{alg:hessian_update}. The iterations of PROXTONE seem to achieve the best of all.

\subsection{Sparse deep learning}

Two kinds of widely used typical deep learning models, which are sparse DNN and CNN, are used to test our method. 

First we trained a deep neural network to classify digits on the MNIST digit recognition benchmark. We used a similar architecture to~\cite{hinton2012improving}. The MNIST~\cite{lecun1998gradient} dataset consists of 28*28 pixel greyscale images of handwritten digits 0-9, with 60,000 training and 10,000 test examples. Our network consisted of: 784 input units, one hidden layer of 1200 units, a second hidden layer of 1200 units, and 10 output units. We ran the experiment using both rectified linear and sigmoidal units. The objective used was the standard softmax regression on the output units. Theano~\cite{bergstra2010theano} was used to implement the model architecture and compute the gradient.

Second we trained a deep convolutional network on CIFAR-10 using max pooling and rectified linear units. The CIFAR-10 dataset~\cite{krizhevsky2009learning} consists of 32*32 color images drawn from 10 classes split into 50,000 train and 10,000 test images. The architecture we used contains two convolutional layers with 48 and 128 units respectively, followed by one fully connected layer of 240 units. This architecture was loosely based on~\cite{goodfellow2013maxout}. Pylearn2~\cite{goodfellow2013pylearn2} and Theano were used to implement the model.

\begin{figure}
\centering
\begin{tabular}{ccc}
\hspace{-5mm}
\begin{tabular}{c}
\includegraphics[width=0.39\linewidth]{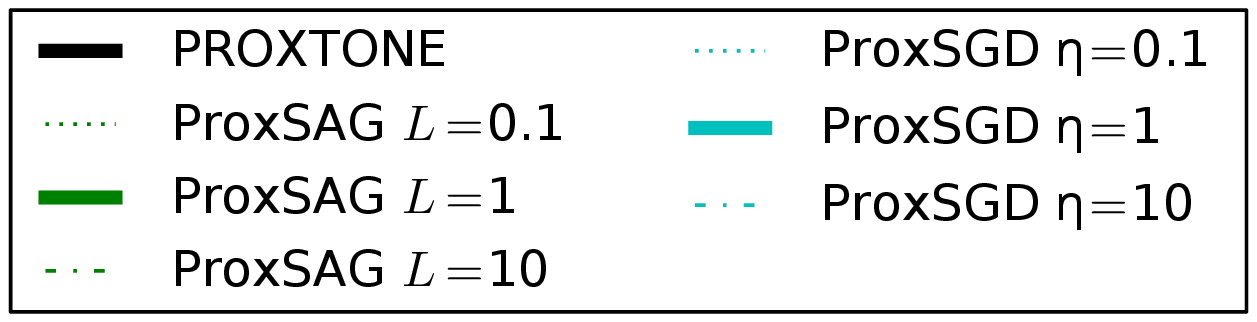} \\
(a)\includegraphics[width=0.39\linewidth]{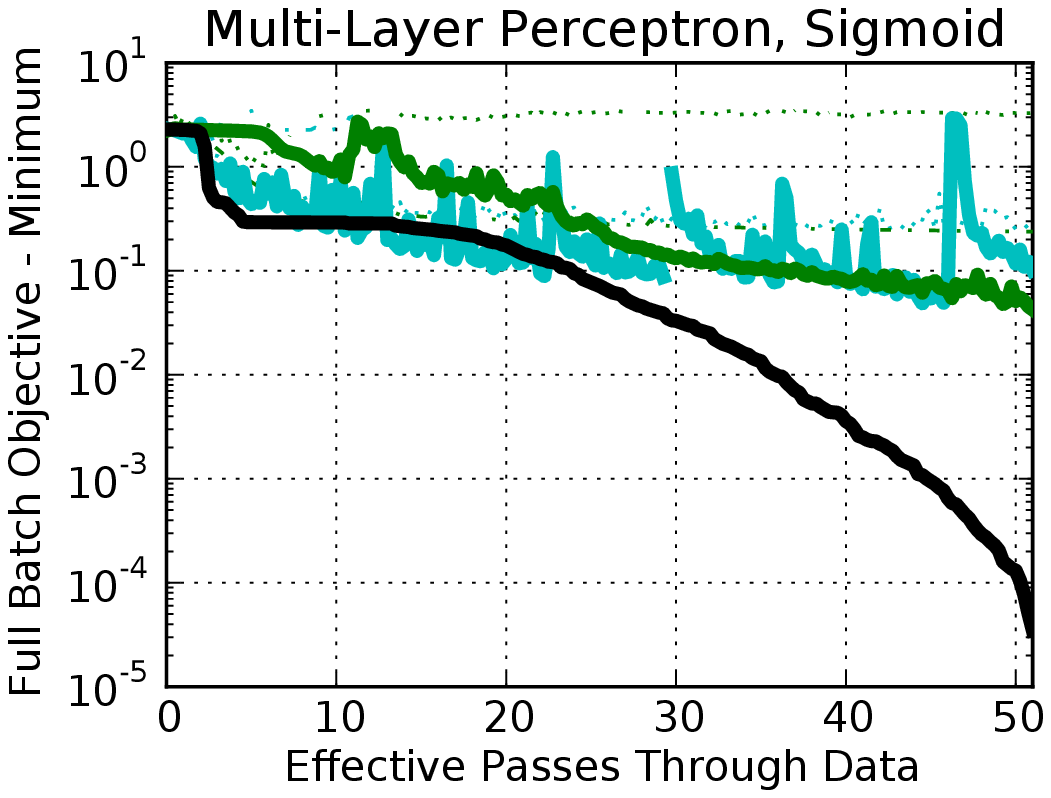}
\end{tabular}
\hspace{-5mm}
 &
\begin{tabular}{c}
\includegraphics[width=0.39\linewidth]{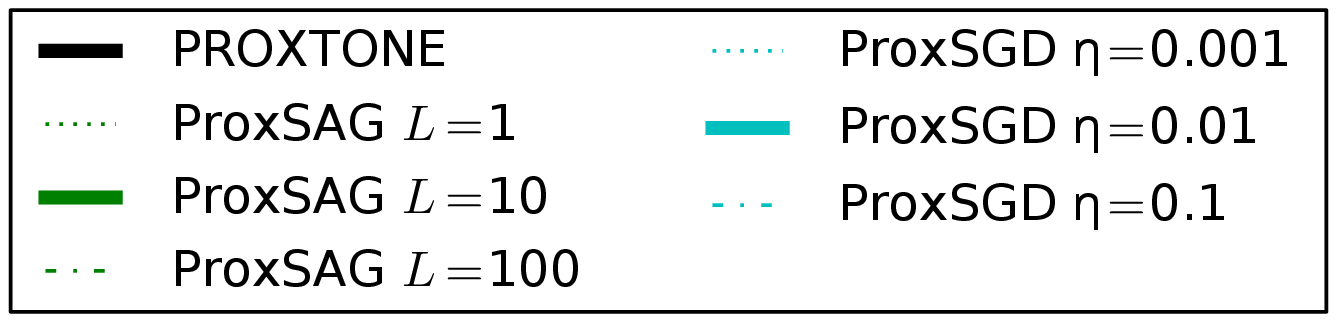} \\
(b)\includegraphics[width=0.39\linewidth]{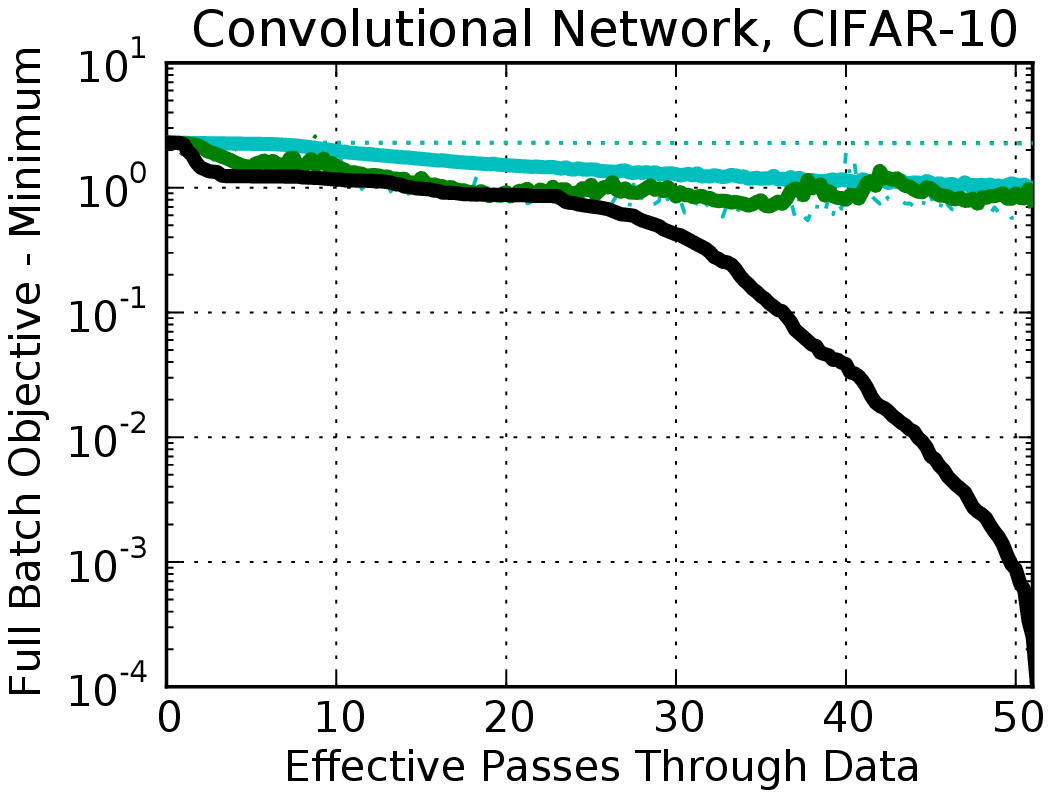}
\end{tabular}
\hspace{-5mm}
\end{tabular}
\caption{
A comparison of PROXTONE to competing optimization techniques for (a) sparse DNN and (b) sparse CNN.
The bold lines indicate the best performing hyperparameter for each optimizer. It can be seen that PROXTONE is free of chosen hyperparameter.
}
\label{hyperparameter_dnn_cnn}
\end{figure}

\begin{figure}
\centering
\hspace{-5mm}
\begin{tabular}{r}
(a)\includegraphics[width=0.8\linewidth]{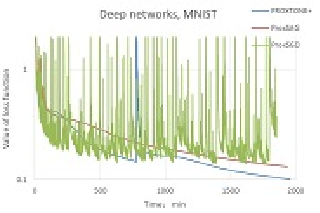} \\
(b)\includegraphics[width=0.8\linewidth]{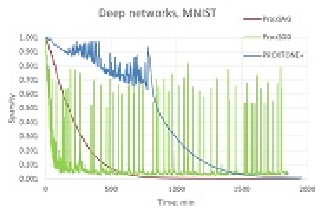} \\
(c)\includegraphics[width=0.8\linewidth]{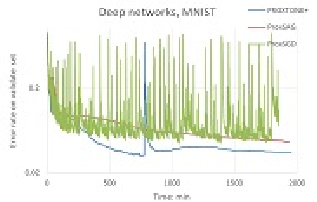}
\end{tabular}
\caption{
A comparison of PROXTONE$^+$ to competing optimization techniques for two objective functions. The objective
functions shown is a multi-layer perceptron with sigmoidal units trained on MNIST digits. (a) value (b) sparsity (c) error.
}
\label{fig results_sparse_dnn}
\end{figure}

\begin{figure}
\centering
\hspace{-5mm}
\begin{tabular}{c}
(a)\includegraphics[width=0.9\linewidth]{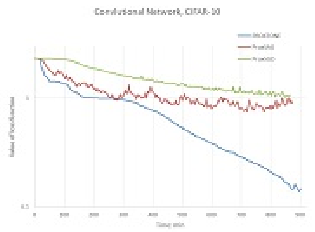} \\
(b)\includegraphics[width=0.9\linewidth]{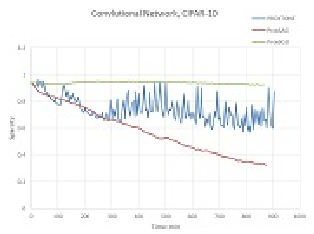}
\end{tabular}
\hspace{-5mm}
\caption{
A comparison of PROXTONE to competing optimization techniques for two objective functions. The objective
functions shown is a multilayer convolutional network with rectified linear units trained on CIFAR-10. (a) value (b) sparsity.
}
\label{fig results_sparse_cnn}
\end{figure}

A preliminary experiment is used to choose the hyperparameter of ProxSAG and ProxSGD for sparse DNN and sparse CNN respectively in Figure~\ref{hyperparameter_dnn_cnn}. Then we do detail measurement of time and sparsity for all the methods.
The Figure~\ref{fig results_sparse_dnn} and~\ref{fig results_sparse_cnn} show that PROXTONE and PROXTONE$^+$ converge nearly twice as fast then the state-of-the-art methods. While for sparsity, PROXTONE$^+$ can reduce the size to about 0.5\% for sparse DNN training. Since there are many share weights in CNN, for sparse CNN training, PROXTONE is much more suitable than PROXTONE$^+$, and reduce the size to about 60\%.

\section{CONCLUSION}
\label{sec:CONCLUSION}

This paper is to make clear the implementation details of PROXTONE and do the numerical evaluations to nonconvex problems, especially sparse deep learning problems. We show that PROXTONE and PROXTONE$^+$ can make full use of gradients, converges much faster than state-of-the-art first order methods in the number of gradients or epochs. It is also showed the methods converges faster also in time, while reduce the size to 0.5\% and 60\% for DNN and CNN models respectively. There are some directions that the current study can be extended. Experiments show that ProxSAG method has good performance, thus it would be meaningful to also make clear the theory for the convergence of ProxSAG~\cite{schmidt2013minimizing}. Second, combine with randomized block coordinate method~\cite{nesterov2012efficiency} for minimizing regularized convex functions with a huge number of varialbes/coordinates. Moreover, due to the trends and needs of big data, we are designing distributed/parallel PROXTONE for real life applications. In a broader context, we believe that the current paper could serve as a basis for examining the method for deep learning on the proximal stochastic methods that employ second order information.

\bibliography{ecai}
\end{document}